# Bayes-Ball: The Rational Pastime
# (for Determining Irrelevance and Requisite Information in Belief Networks and Influence Diagrams)

Ross D. Shachter
Engineering-Economic Systems and Operations Research Dept.
Stanford University
Stanford, CA 94305-4023
shachter@stanford.edu

## Abstract

One of the benefits of belief networks and influence diagrams is that so much knowledge is captured in the graphical structure. In particular, statements of conditional irrelevance (or independence) can be verified in time linear in the size of the graph. To resolve a particular inference query or decision problem, only some of the possible states and probability distributions must be specified, the "requisite information."

This paper presents a new, simple, and efficient "Bayes-ball" algorithm which is well-suited to both new students of belief networks and state of the art implementations. The Bayes-ball algorithm determines irrelevant sets and requisite information more efficiently than existing methods, and is linear in the size of the graph for belief networks and influence diagrams.

## 1 Introduction

One of the benefits of belief networks and influence diagrams is that so much knowledge is captured in the graphical structure. Without the need to compute with or even assess possible states and probabilities, many important properties of a model can be recognized (Pearl 1988; Shachter 1988). To fully specify a network, the possible states and probability distributions must be obtained for all variables. For a particular inference query or decision problem, only some of that information is requisite.

Because this information is stored in the network structure, it lends itself to efficient analysis. In particular, statements of conditional irrelevance (or independence) in belief networks can be verified in time linear in the size of the graph (Geiger et al 1989). Identification of requisite information can also be determined in time linear in the size of the graph. These algorithms have been generalized to deal with deterministic nodes in belief networks and influence diagrams (Geiger et al 1990; Shachter 1990).

This paper introduces the Bayes-Ball algorithm, a simpler and more efficient algorithm to identify conditional irrelevance and requisite information. For belief networks, Bayes-Ball runs in time linear in the size of the active part of the graph, so it is considerably faster when most of a graphical knowledge base is irrelevant. It also corrects an error in the requisite information algorithm given in Geiger (1990). More significantly, for decision problems it runs in time linear in the size of the graph; up until now the fastest algorithm (Shachter 1990) has been O((number of decisions)(graph size)). Finally, the decision algorithm has been extended to allow multiple separable value nodes.

The identification of irrelevant nodes and requisite information is a fundamental operation in any belief network or influence diagram processor. It allows quick pruning of graphical knowledge bases, provides simple explanations for fundamental relationships (Poh and Horvitz 1996), and even suggests new algorithms (Lin and Drudzel 1997). Although it has not been published before now, the Bayes-ball algorithm for belief networks has been freely shared in the community and is implemented in systems such as Knowledge Industries' DXPress$^{TM}$ and Microsoft's MSBN$^{TM}$.

Section 2 contains the notation and framework for the paper and the previous results from the literature. Section 3 develops and proves the Bayes-ball algorithm for belief networks, and Section 4 extends and refines it for influence diagrams.

## 2 Notation and Framework

### 2.1 Irrelevance

A *structured belief network* $B = (N, A, F)$ consists of nodes $N$ and directed arcs $A$ which together form a directed acyclic graph $G(N, A)$, and a subset $F$ of the nodes that are deterministically (functionally) re-lated to their parents. Corresponding to each node $j$ is an <u>uncertain variable</u> $X_j$. Consider the belief network shown in Figure 1a. This represents a contest in which the prize depends on the state of two flipped coins. *Coin 1* and *Coin 2* are oval <u>probabilistic nodes</u> and *Win Prize* is a double oval <u>deterministic node</u>.

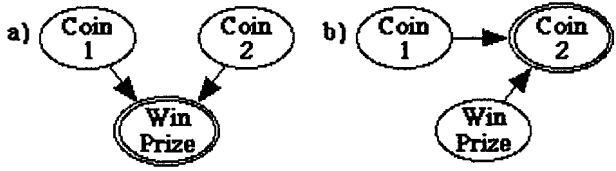

Figure 1: Different assessment orders reveal different irrelevances

In referring to the graph $G(N, A)$, for any node $i \in N$, $Pa(i)$ denotes the parents of $i$ and $De(i)$ denotes the descendants of $i$.

A joint probability distribution over $X_N$ is said to admit a <u>directed factorization</u> with respect to a structured belief network $B = (N, A, F)$ if $X_j$ is a deterministic function of $X_{Pa(j)}$ for all $j \in F$ and

$$\Pr\{X_N\} = \prod_{j \in N} \Pr\{X_j | X_{Pa(j)}\}.$$

Given a joint probability distribution for $X_N$, $\Pr\{X_N\}$, $X_J$ is <u>probabilistically irrelevant</u> (usually called "independent") to $X_L$ given $X_K$ for some $J, K, L \subseteq N$ if

$$\Pr\{X_J | X_K, X_L\} = \Pr\{X_J | X_K\}.$$

Given a structured belief network $B = (N, A, F)$ and sets $J, K, L \subseteq N$, $X_J$ is said to be <u>irrelevant</u> to $X_L$ given $X_K$ in $B$, denoted $X_J \perp_B X_L | X_K$, if $X_J$ is probabilistically irrelevant to $X_L$ given $X_K$ for any joint probability distribution $\Pr\{X_N\}$ that admits a directed factorization with respect to $B$.

Less formally, $X_J$ is irrelevant to $X_L$ given $X_K$ in $B$ if, having observed $X_K$, one can learn nothing more about $X_J$ by also observing $X_L$. The following proposition is well known (Pearl 1988).

**Proposition 1 (Irrelevance Decomposition)**
*Given any structured belief network $B = (N, A, F)$ and any subsets $J, K, L \subseteq N$, $X_J \perp_B X_L | X_K$ if and only if $X_j \perp_B X_l | X_K$ for all $j \in J$ and $l \in L$.*

It is important to understand that probabilistic irrelevance is not the same as the irrelevance represented by a belief network. For example, Proposition 1 does not hold for particular probability distributions unless their independence is completely representable in a structured belief network. Consider the coin flipping contest shown in Figure 1a. The decision-maker believes that the two flips are irrelevant but that the prize is determined by them. If he believes that the coins are both fair and the prize will be awarded if the two coins match, then the network in Figure 1b is also valid, since he believes that the state of one coin tells him nothing about whether he will win. (That would not be true if the coins were not fair or the contest were won with two heads.) A misapplication of Proposition 1 would suggest that seeing both coins would tell him nothing about whether he will win!

### 2.2 Deterministic Irrelevance

Given any structured belief network $B = (N, A, F)$ and a deterministic node $j \in F$, the variable $X_j$ is certain or <u>effectively observed</u> if the parents of node $j$, $Pa(j)$, are observed. Similarly, if each parent of $j \in F$ is either observed or effectively observed, then the variable $X_j$ is also effectively observed. More formally, if the variables $X_K$ are observed for some $K \subseteq N$, a node $j$ is said to be <u>functionally determined</u> by $K$ if either $j \in K$ or $X_j$ is a deterministic function of $X_K$. The set of nodes functionally determined by $K$, $F_K$, can be described by the recursive equation,

$$F_K \leftarrow K \cup \{i \in F : Pa(i) \subseteq F_K\},$$

corresponding to nested instances of effective observation. For example, in Figure 1a, if the two coins are observed then it is functionally determined whether the prize will be awarded.

**Proposition 2 (Deterministic Irrelevance)**
*Given a structured belief network $B = (N, A, F)$ and sets $J, K \subseteq N$, $X_J$ is functionally determined by $X_K$ in belief network $B$ if and only if $X_J \perp_B X_N | X_K$.*

As a special case of Proposition 2,

$$X_{F_K} \perp_B X_N | X_K \text{ for all } K \subseteq N.$$

### 2.3 Target Sets

Efficient computation of the following sets is one objective of this paper. The other objective is efficient computation of the related requisite sets for sequential decision problems represented by influence diagrams.

The <u>irrelevant nodes</u> for $X_J$ given $X_K$, denoted $X_{N_i}(J|K)$, are those nodes which are conditionally irrelevant to $X_J$ given $X_K$,

$$N_i(J|K) = \{i \in N : X_i \perp_B X_J | X_K\}.$$

The <u>requisite</u> or required information needed to compute $\overline{\Pr\{X_J | X_K\}}$ depends on the probabilistic irrelevance, rather than the conditional irrelevance revealed in the belief network graph. Thus, from the graph we can only recognize the information which "might" be needed for any probability distribution.

The <u>requisite probability nodes</u> for $J$ given $K$, denoted $N_p(J|K)$, are those nodes for which conditional probability distributions (and possible states) might be needed to compute $\Pr\{X_J | X_K\}$. The conditional probability distribution for a deterministic node is usually specified as a deterministic function.

The *requisite observations* for $J$ given $K$, $N_e(J|K) \subseteq K$, are those observed nodes for which observations (and hence the possible states which might be observed) might be needed to compute $\Pr\{X_J|X_K\}$.

In previous papers (Shachter 1988; Shachter 1990), these sets were referred to as the "minimal sets of relevant nodes," $N_\pi(J|K) = N_p(J|K)$ and $N_\Omega(J|K) = N_p(J|K) \cup N_e(J|K)$.

### 2.4 D-separation

The Bayes-ball algorithm is strongly based on the concept of d-separation (Geiger et al 1989; Pearl 1988; Pearl et al 1990) and its deterministic generalization, D-separation (Geiger et al 1990).

An *active path* from $J$ to $L$ given $K$ is a simple trail (or undirected path) between $i \in L$ and $j \in J$, such that every node with two incoming arcs on the trail is or has a descendant in $K$; and every other node on the trail is not functionally determined by $K$. As a special case, active paths can be just a single node, that is, $i = j$.

Given sets of nodes, $J$, $K$, and $L$ from belief network $B$, $K$ is said to *D-separate* $J$ from $L$ in $B$ if there is no active path from $J$ to $L$ given $K$. This condition determines all irrelevancies represented in the belief network (1990).

**Theorem 1 (D-Separation)** *Given a structured belief network $B = (N, A, F)$ and $J, K, L \subseteq N$, $X_J \perp_B X_L | X_K$ if and only if $K$ D-separates $J$ from $L$ in $B$.*

Geiger (1989; 1990) devised linear-time reachability algorithms for finding active paths from $J$ given $K$ that inspired the Bayes-ball algorithm. Unfortunately, active paths are not adequate for gathering requisite information since they cannot enter clusters of functionally determined nodes. The active path algorithms also must search the entire graphical model to find the ancestors of $K$ and the nodes functionally determined by $K$.

An alternative characterization of the irrelevance represented by a belief network involves a conversion to related undirected graphical structures (Lauritzen et al 1990). Those results could be generalized for determinacy following the approach in (Shachter 1991) but, due to the nature of the conversion, the undirected approaches cannot find all nodes irrelevant to $J$ given $K$ in linear time.

## 3 The Bayes-Ball Algorithm

This section presents two versions of the Bayes-ball algorithm, a simpler, preliminary version and a more refined, final version. The final version determines the irrelevant and requisite sets for a given inference query in time linear in the size of the belief network.

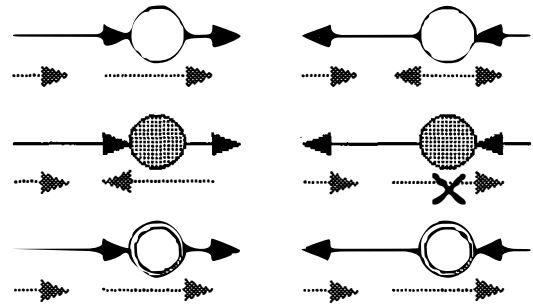

Figure 2: The Bayes-ball bounces back and passes through differently based on the type of node and the direction from which it came

### 3.1 Simpler Bayes-Ball

The simpler version of the Bayes-ball algorithm for $J$ given $K$ sends a bouncing ball to visit nodes in the belief network, starting from nodes $J$. Depending on the type of node and the direction from which the ball came to visit (from parents; from children), the ball can *pass through* the node (from any parent to all children; from any child to all parents), *bounce back* (from any parent to all parents; from any child to all children), or be *blocked*. This is summarized in Figure 2 in which:

- an unobserved probabilistic node passes balls through but also bounces back balls from children;
- an observed node bounces back balls from parents but blocks balls from children; and
- a deterministic unobserved node always passes balls through.

**Algorithm 1 (Simpler Bayes-Ball Algorithm)**
*The algorithm explores a structured belief network $B = (N, A, F)$ with respect to the expression $\Pr\{X_J|X_K\}$.*

1. *Visit each node in $J$ (as if from one of its children).*

2. *When visiting a node $j$:*

   (a) *If the visit to $j$ is from a child:*
      i. *If $j \in K$ then do nothing;*
      ii. *otherwise, if $j \in F$ then visit each of $j$'s parents;*
      iii. *otherwise, visit each of $j$'s parents and each of $j$'s children.*
   (b) *If the visit to $j$ is from a parent:*
      i. *If $j \in K$ then visit each of $j$'s parents;*
      ii. *otherwise, visit each of $j$'s children.*

As an example of the simpler Bayes-ball algorithm, consider the belief network shown in Figure 3a, in which $J = \{6\}$ (denoted by the parents of the "ghost" rounded rectangle) and $K = \{2, 5\}$ (denoted by shading). The simpler Bayes-ball algorithm is illustrated in

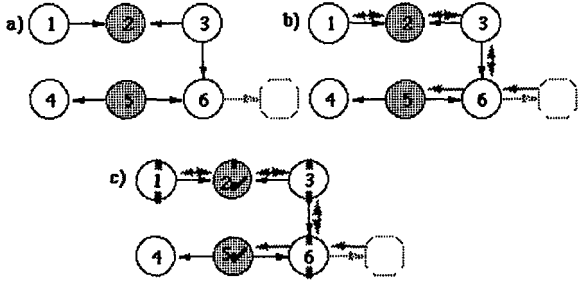

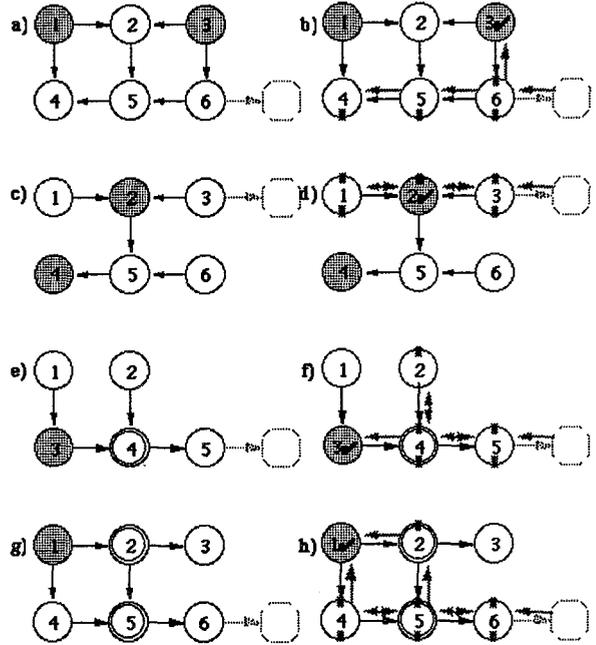

Figure 3: The simpler and refined versions of the Bayes-ball algorithm applied to a small belief network.

Figure 3b. Starting from the rounded rectangle, visit its parent, node 6. From node 6 pass the ball through to nodes 3 and 5 (and to the children of node 6 if it had any). Node 5 does not send the ball anywhere, but node 3 bounces it to its children, nodes 2 and 6, (and would pass it through to its parents if it had any). Node 6 has no children to pass to, but node 2 bounces the ball back to its parents, nodes 1 and 3. In turn, nodes 1 and node 3 both bounce it back to node 2. Although the ball will keep bouncing forever, at this point it has visited all of the nodes it will ever visit from all of the directions it will ever visit them.

### 3.2 The Final Bayes-Ball Algorithm

An obvious improvement to this algorithm is to maintain a list of nodes to be visited from parents and from children. Another, more critical improvement is to only visit the same arc in the same direction once. (This is not only more efficient, but necessary to terminate the infinite loops.) It is accomplished by marking the top of a node when the ball is sent to the node's parents, marking the bottom of the node when the ball is sent to the node's children, and checking an observed node when it is visited. These marks not only allow the algorithm to terminate, but they also record significant results. Returning to the belief network shown in Figure 3a, apply these modifications to obtain the network shown in Figure 3c. At the start, node 6 is visited (as if from its child) and it sends the ball to its parents and children, marking its top and bottom. Node 5 does not send the ball anywhere, so it is checked but not marked. Now node 3 is visited from its child node 6, so it sends the ball to its parents and children and marks its top and bottom. Node 6 receives the ball from parent node 3 and it would send it to its children if its bottom were not marked already. Node 2 receives the ball from parent node 3, sends it to its parents, and marks its top. Node 1 receives the ball from child node 2, sends it to its parents and children, and marks its top and bottom. Finally node 2 and node 3 receive the ball and recognize that there is nothing new to do with it.

The Bayes-ball algorithm was run on the belief network queries shown in parts a, c, e, and g of Figure 4

Figure 4: The Bayes-ball algorithm applied to some small belief networks

to obtain the networks shown in parts b, d, f, and h. In Figure 4b, nodes 1 and 2 are not visited. In Figure 4d, nodes 4, 5, and 6 are not visited because observed node 2 will not let the ball pass through, although node 2 does bounce it back to its parents. In Figure 4f, the ball is passed through nodes 5 and 4, but it is not passed through node 3, so node 1 is not visited. Finally, in Figure 4h, the ball is passed through nodes 6, 5, 4, and 2, but because there is no bounce back from node 1, the deterministic node 2 never passes the ball through to node 3.

These examples help to illustrate some of the properties of the Bayes-ball algorithm:

- the node $j$ is visited by the algorithm if and only if observing $x_j$ might change one's belief about $X_J$ given $X_{K \setminus \{j\}}$;
- we need no information about any node which has not been visited;
- we might need to know what state was observed for any observed node which is visited;
- the ball bounces back to a parent from a child only if that child is observed, such as node 2 in Figure 4d;
- the ball never bounces back to a child from a parent functionally determined by $K$, such as nodes 1 and 2 in Figure 4h;
- the ball is passed to parents to find requisite information and passed to children looking for relevant observations;
- any node not marked on its bottom is irrelevant to $J$ given $K$;

- we might need a conditional probability distribution (or deterministic function) for any node marked on its top.

**Algorithm 2 (Bayes-Ball)** *The algorithm explores a structured belief network $B = (N, A, F)$ with respect to the expression $\Pr\{X_J | X_K\}$ and constructs the sets of relevant and requisite nodes.*

1. Initialize all nodes as neither visited, nor marked on the top, nor marked on the bottom.

2. Create a schedule of nodes to be visited, initialized with each node in $J$ to be visited as if from one of its children.

3. While there are still nodes scheduled to be visited:
   (a) Pick any node $j$ scheduled to be visited and remove it from the schedule. Either $j$ was scheduled for a visit from a parent, a visit from a child, or both.
   (b) Mark $j$ as visited.
   (c) If $j \notin K$ and the visit to $j$ is from a child:
       i. if the top of $j$ is not marked, then mark its top and schedule each of its parents to be visited;
       ii. if $j \notin F$ and the bottom of $j$ is not marked, then mark its bottom and schedule each of its children to be visited.
   (d) If the visit to $j$ is from a parent:
       i. If $j \in K$ and the top of $j$ is not marked, then mark its top and schedule each of its parents to be visited;
       ii. if $j \notin K$ and the bottom of $j$ is not marked, then mark its bottom and schedule each of its children to be visited.

4. The irrelevant nodes, $N_i(J|K)$, are those nodes <u>not</u> marked on the bottom.

5. The requisite probability nodes, $N_p(J|K)$, are those nodes marked on top.

6. The requisite observation nodes, $N_e(J|K)$, are those nodes in $K$ marked as visited.

### 3.3 Bayes-Ball Proofs

The section proves that Bayes-ball properly computes the irrelevant and requisite sets and does it in sublinear time.

**Theorem 2 (Bayes-Ball Irrelevance)** *Given a structured belief network $B = (N, A, F)$ and $J, K, L \subseteq N$, $X_J \perp_B X_L | X_K$ if and only if $L \subseteq N_i(J|K)$, as determined by the Bayes-Ball algorithm.*

**Proof:** Based on Theorem 1 it is sufficient to show that a node $l$ will be marked on the bottom if and only if there is an active path from $J$ to $l$ given $K$.

First, any node $i$ is or has a descendant in $K$ if and only if it would always bounce the ball back to its parents after one of them sent the ball to it, since the bounce back can only come from an observed descendant.

Second, $i$ is functionally determined by $K$ if and only if it would never bounce the ball back to its children after one of them sent the ball to it, since any unobserved probabilistic ancestor would bounce it back.

Therefore, given any active path from $J$ to $l$, Bayes-ball can travel the path (perhaps with the detours just described) and it will be visiting the children of node $l$, either by passing through or bouncing back. Similarly, if the Bayes-ball is sent from $l$ to its children then there must be an active path from $J$ to $l$. □

**Theorem 3 (Bayes-Ball Requisite Sets)** *Given a structured belief network $B = (N, A, F)$, for any $J, K \subseteq N$, the Bayes-ball algorithm determines the requisite probability nodes, $N_p(J|K)$, and the requisite observation nodes, $N_e(J|K)$.*

**Proof:** Whether a node $j$ is visited does not depend at all on whether it is observed or deterministic, but only on the properties of other nodes leading to it. If the algorithm visits an unobserved probabilistic node $j$, then it will mark $j$ on the bottom and, by the preceding theorem, $j$ appears relevant to $J$ given $K$. Thus, the algorithm will visit node $j$ if and only if, keeping all other nodes unchanged but changing $j$ to be unobserved and probabilistic, $X_j$ might be relevant to $X_J$ given $X_K$.

As a result, the state of an observed node $k$ provides no information about $J$ given $K$ unless it is visited. Thus, $N_e(J|K)$ comprises those nodes in $K$ which are visited during the algorithm.

To determine $N_p(J|K)$, add a new observed parent (a probability distribution/deterministic function) to every node. These new nodes would not have requisite observations unless they were visited, and this would occur only when their corresponding children have been marked on top. □

The last part of the preceding proof relates to the flaw in the requisite information algorithm in Geiger (1990). In that paper, relevance is used to recognize requisite distributions, but that can introduce uncertainty and eliminate functional determinacy. On the other hand, in this work, requisite information is based on the need for observations and thus no spurious uncertainty is introduced. Bayes-ball is able to gather both irrelevance and requisite information by recognizing their distinction—visiting parents for requisite information and visiting children for relevance.

**Theorem 4 (Bayes-ball Complexity)**
*Given a structured belief network $B = (N, A, F)$, for any $J, K \subseteq N$, the Bayes-ball algorithm runs in $O(|N| + |A_v|)$, where $A_v$ are the arcs incident to the nodes marked during the algorithm. In the worst case,*

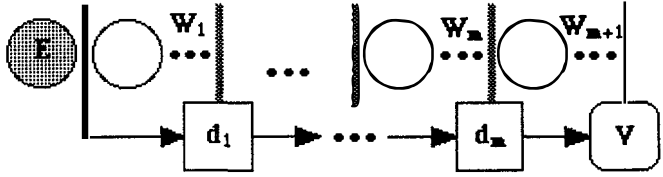

Figure 5: Informational arcs, separable values, and decision windows

it is linear time in the size of the graph.

**Proof:** Each node has to be visited to initialize flags and to harvest results and then at most once for each "active" incident arc. Only the active arcs will be visited, because the algorithm searches locally to determine whether a node has an observed descendant or is functionally determined. □

## 4  Bayes-Ball for Decisions

The results in Section 3 determining requisite sets for probabilistic inference can be applied to problems of decision making under uncertainty (Shachter 1986; Shachter 1988). In this section, the Bayes-ball algorithm is adapted to determine the requisite information for decision problems represented by influence diagrams, yielding a new, significantly more efficient and powerful algorithm.

The *influence diagram* is a belief network augmented with rectangular *decision nodes,* representing variables under the control of the decision maker, and rounded rectangular *value nodes,* representing the criterion whose expectation the decision maker maximizes to choose among decision alternatives. The decisions, $D$, are assumed to be ordered in time, $d_1, \ldots, d_m$, and the uncertain variables are partitioned into sets $E, W_1, \ldots, W_{m+1}$ such that the uncertain variables $W_i$ are assumed to be observed before decision $d_i$ but after decision $d_{i-1}$. Some of the uncertainties, $W_{m+1}$, will not be observed before any of the decisions, while others, $E$, have already been observed. We assume that if there are multiple value nodes in $V$ that they are related by a *separable function,* that is, either a sum or a product (of nonnegative factors). This structure is illustrated in Figure 5. Implicit in the figure is the *no forgetting* assumption that at the time of any decision, the decision maker will remember all previous decisions and the uncertain variables known at the time they were made. These are captured in the *information sets,* $I(d_i)$, for each decision $d_i$. It follows that $I(d_1) = W_1 \cup E$ and $I(d_i) = W_i \cup d_{i-1} \cup I(d_{i-1})$ for $i = 2, \ldots, m$. These information sets are indicated in the influence diagram by *information arcs* into the decision nodes.

Consider the influence diagram shown in Figure 6a modeling the design of an experiment. The decision maker seeks to maximize expected *Benefit* minus *Cost*.

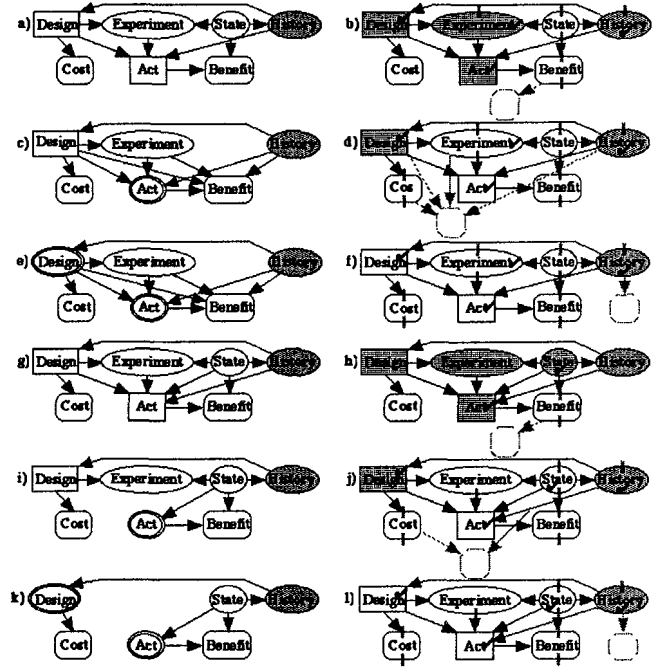

Figure 6: A design of experiment decision model with different informational assumptions

*Benefit* depends on the uncertain variable *State* and the decision *Act*. He has already observed *History* which depends on *State*, and before making the *Act* decision, he will observe *Experiment*, which depends on both the *Design* decision and the *State*. There is a *Cost* to the *Design* decision. This problem is solved recursively through dynamic programming to determine an optimal policy, $d_m^*(x_{I(d_m)})$ for the latest decision, $d_m$ as a function of the information available at the time of the decision. Only the value nodes which are descendants of $d_m$ are affected by this policy, $V_m = V \cap De(d_m)$ (Heckerman and Shachter 1995; Shachter and Peot 1992). This policy must satisfy

$$\mathrm{E}\{V_m | d_m^*(x_{I(d_m)}), x_{I(d_m)}\} = max_d \mathrm{E}\{V_m | d, x_{I(d_m)}\}.$$

This suggests the use of the Bayes-ball algorithm to determine the requisite sets. The decision $d_m$ can be replaced by the optimal policy to obtain the influence diagram shown in Figure 6c, in which *Act* is now a deterministic function of the requisite observations, *Design, Experiment,* and *History*. At this point, the value nodes $V_m$ are irrelevant of any earlier decisions given the requisite observations at the time of decision $d_m$. So the value for $d_{m-1}$ can be characterized in terms of $V_{m-1}$ and the requisite observations from the first step. Continuing with the example, *Cost, Design, Experiment,* and *History* are the goals for *Design,* and the policy is a function of *History* as shown in Figure 6e.

Suppose instead that the decision-maker were able to observe the *State* before the *Act* decision, as shown in Figure 6g. Now, *State* is the only requisite observed node before *Act*, yielding the diagram shown in Fig-ure 6i. As a result, *Cost* and *State* are the goals for *Design* as shown in Figure 6k.

The requisite probability nodes for decision $i$, denoted $N_p^i$, are those nodes for which conditional probability distributions (and possible states) might be needed to evaluate the decision problem starting with decision $i$. $N_p^0$ is the set of requisite probability nodes for the entire decision problem.

The *requisite observations* for decision $i$, denoted $N_e^i$, are those nodes which can be observed before decision $i$ that might be worth observing (and hence for which the possible states might be needed) in order to evaluate the decision problem starting with decision $i$. $N_e^0$ is the set of requisite observation nodes before the entire decision problem, assuming that we have observed nodes $E$ now; if $E = \emptyset$ then $N_e^0 = \emptyset$.

**Algorithm 3 (Decision Bayes-Ball)**
*This algorithm constructs the sets of requisite nodes for an influence diagram with separable value nodes $V$, evidence nodes $E$, and time-ordered decision nodes $D = \{d_1, \ldots, d_m\}$.*

1. *start with the last decision, $d_m$:*

   (a) *Determine the relevant value nodes, $V_m = V \cap De(d_m)$. (If there is only one value node for the problem, it should be in $V_m$.)*

   (b) *Run the Bayes-ball algorithm on $V_m | \{d_m\} \cup I(d_m)$, ignoring any informational arcs.*

   (c) *If $d_m$ is not marked as visited then decision $d_m$ is irrelevant–it has no effect on the decision-maker's value.*

   (d) *The requisite observation nodes for decision $d_m$ are the nodes in $I(d_m)$ marked as visited,*
   $$N_e^m = N_e(V_m | \{d_m\} \cup I(d_m)) \cap I(d_m).$$

   (e) *The requisite probability nodes starting with decision $d_m$ are the nodes marked on top,*
   $$N_p^m = N_p(V_m | \{d_m\} \cup I(d_m)).$$

2. *iterate backwards for each earlier decision $d_i$, $i = m - 1, \ldots, 1$:*

   (a) *Determine the relevant value nodes, $V_i = V \cap (De(d_i) \setminus De(d_{i+1}))$, ignoring any informational arcs in determining descendants.*

   (b) *Resume the Bayes-ball algorithm on $V_i \cup N_e^{i+1} | \{d_i\} \cup I(d_i)$, ignoring any informational arcs.*

   (c) *If $d_i$ is not marked as visited then decision $d_i$ is irrelevant.*

   (d) *The requisite observation nodes for decision $d_i$, $N_e^i$, are those nodes in $I(d_i)$ marked as visited.*

   (e) *The requisite probability nodes starting with decision $d_i$, $N_p^i$, are all of the nodes marked on top.*

3. *compute the requisite information at the start of the decision problem (now) given current observations $E$:*

   (a) *Resume the Bayes-ball algorithm on $N_e^1 | E$, ignoring any informational arcs.*

   (b) *The requisite observations needed now, $N_e^0$, are those nodes in $E$ marked as visited. (If $E$ is empty, then no observations are needed now.)*

   (c) *The requisite probability nodes starting now, $N_p^0$, are all of the nodes marked on top.*

Consider the influence diagram examples shown in Figure 6. As explained earlier, the left column represents the influence diagram at each step as decisions are replaced by policies for the situations without and with observation of *State* before the *Act* decision. The right column shows how the Bayes-ball algorithm for decisions is applied sequentially to the original problem structure. Bayes-ball works efficiently on influence diagrams by resuming rather than restarting at each step. The sets computed by the algorithm are shown in the following table:

| Figure 6 | $i$ | $N_e^i$ | $N_p^i$ |
|---|---|---|---|
| a/b | 2 | Design, History, Experiment | Benefit, History, Experiment, State |
| c/d | 1 | History | above + *Cost* |
| e/f | 0 | History | above |
| g/h | 2 | State | Benefit |
| i/j | 1 | History | Benefit, Cost, History, State |
| k/l | 0 | History | above |

**Theorem 5 (Decision Bayes-Ball)** *Given an influence diagram with separable value nodes and evidence now, Algorithm 3 will determine the requisite node sets in $O(|N| + |A|)$, that is, linear time in the number of nodes and arcs.*

**Proof:** The correctness of the requisite sets follows from the Bayes-ball algorithm and the decision theory summarized in this section, so the big question here is the computational complexity.

There are two issues to be resolved there. First, the sets of value nodes can be determined in linear time in the size of the graph. Second, the Bayes-ball algorithm for each decision can be resumed during the algorithm, keeping the previous results. This works because the information sets are monotone, $E \subseteq \{d_1\} \cup I(d_1) \subseteq \ldots \subseteq \{d_m\} \cup I(d_m)$. Once the algorithm has been run on $V_i \cup N_e^{i+1} | \{d_i\} \cup I(d_i)$, it is easy to run it on $V_{i-1} \cup N_e^i | \{d_{i-1}\} \cup I(d_{i-1})$. The nodes $(\{d_i\} \cup I(d_i) \setminus \{d_{i-1}\} \cup I(d_{i-1})) = \{d_i\} \cup W_i$ are no longer observed, but the previously requisite observations $N_e^{i+1} \setminus N_e^i$ are now scheduled to receive messages from their children, so they will now pass the Bayes-ball through to their parents automatically! The Bayes-ball sets of nodes visited, marked on the top, and marked on the bottom, can grow monotonically as the decisions are visited in reverse order. The computation of $N_e^i$ is still correct, since only the currently observed visited nodes are included and the number of node visits is $O(|N| + |A|)$. □

## 5 Conclusions

This paper has introduced a simple and efficient algorithm to compute irrelevant and requisite sets for inference and decision problems represented as structured belief networks and influence diagrams. Because these procedures can be used before states and distributions are assessed they are quite useful. Although the efficiency gain over earlier belief network algorithms is modest (sub linear vs. linear time), the computation of requisite information is performed at the same time as irrelevance is determined.

The significant improvement presented here is in determining requisite information for decision problems. The new algorithm is linear time instead of O((number of decisions)(graph size)), and can exploit separable values.

These algorithms recognize the special properties of deterministic relationships. Such models are becoming increasingly useful as new developments arise in causal models (Heckerman and Shachter 1995). An interesting extension of the deterministic model would be to represent determinism in more than one assessment order, such as representing when deterministic relationships are invertible.

Another extension is to apply Bayes-ball to cyclical networks (Pearl and Dechter 1996). Bayes-ball can be applied to such networks without any modifications.

Although the algorithm recognizes the full information requirements to value decision problems, it can be modified to recognize some additional efficiencies in the dimensions of policy. For example, in Figure 6k, the choice of *Design* would not depend on *History* if the total value were the sum of *Cost* and *Benefit*, although the value derived would indeed depend on *History*. As a result we wouldn't need a probability distribution for *History* and *State* to determine the optimal policy, but we would need them to value that policy.

### Acknowledgements

This paper has benefited from the comments, suggestions, and ideas of many friends and students, most notably Danny Geiger, Derek Ayers, Mark Peot, David Heckerman, Marcia Tsugawa, Robert Korsan, Richard Barlow, and anonymous referees.

### References

Geiger, D., T. Verma, and J. Pearl. "d-separation: from theorems to algorithms." In **Fifth Workshop on Uncertainty in Artificial Intelligence in University of Windsor, Ontario**, 118-125, 1989.

Geiger, D., T. Verma, and J. Pearl. "Identifying independence in Bayesian networks." **Networks** 20 (1990): 507-534.

Heckerman, D. and R. Shachter. "Decision-Theoretic Foundations for Causal Reasoning." **Journal of Artificial Intelligence Research** 3 (1995): 405-430.

Lauritzen, S. L., A. P. Dawid, B. N. Larsen, and H.-G. Leimer. "Independence properties of directed Markov fields." **Networks** 20 (1990): 491-505.

Lin, Y. and M. J. Drudzel. "Computational Advantages of Relevance Reasoning in Bayesian Belief Networks." In **Uncertainty in Artificial Intelligence: Proceedings of the Thirteenth Conference**, 342-350. San Mateo, CA: Morgan Kaufmann, 1997.

Pearl, J. **Probabilistic Reasoning in Intelligent Systems**. San Mateo, CA: Morgan Kaufmann, 1988.

Pearl, J. and R. Dechter. "Identifying Independencies in Causal Graphs with Feedback." In **Uncertainty in Artificial Intelligence: Proceedings of the Twelfth Conference**, 420-426. San Mateo, CA: Morgan Kaufmann, 1996.

Pearl, J., D. Geiger, and T. Verma. "The Logic of Influence Diagrams." In **Influence Diagrams, Belief Nets, and Decision Analysis**, eds. R. M. Oliver and J. Q. Smith. 67-87. Chichester: Wiley, 1990.

Poh, K. L. and E. Horvitz. "A Graph-Theoretic Analysis of Information Value." In **Uncertainty in Artificial Intelligence: Proceedings of the Twelfth Conference**, 427-435. San Mateo, CA: Morgan Kaufmann, 1996.

Shachter, R. "A Graph-Based Inference Method for Conditional Independence." In **Uncertainty in Artificial Intelligence: Proceedings of the Seventh Conference**, eds. B D'Ambrosio, P Smets, and P Bonissone. 353-360. San Mateo, CA: Morgan Kaufmann, 1991.

Shachter, R. D. "Evaluating Influence Diagrams." **Ops. Rsrch.** 34 (November-December 1986): 871-882.

Shachter, R. D. "Probabilistic Inference and Influence Diagrams." **Ops. Rsrch.** 36 (July-August 1988): 589-605.

Shachter, R. D. "An Ordered Examination of Influence Diagrams." **Networks** 20 (1990): 535-563.

Shachter, R. D. and M. A. Peot. "Decision Making Using Probabilistic Inference Methods." In **Uncertainty in Artificial Intelligence: Proceedings of the Eighth Conference**, 276-283. San Mateo, CA: Morgan Kaufmann, 1992.